\documentclass[letterpaper, 10 pt, conference]{ieeeconf}  

\IEEEoverridecommandlockouts                              

\usepackage{cite}
\usepackage{amsmath,amssymb,amsfonts}
\usepackage{graphicx}
\usepackage{algorithmic}
\usepackage{textcomp}
\usepackage{xcolor}
\usepackage{booktabs}
\usepackage{multirow}
\usepackage{makecell}
\usepackage{threeparttable}
\usepackage{nameref}
\usepackage{hyperref}

\title{\LARGE \bf
IDEA: Insensitive to Dynamics Mismatch via Effect Alignment for \\ Sim-to-Real Transfer in Multi-Agent Control
}

\author{
Chenlong Liu$^{1,\dagger}$, Zhuohui Zhang$^{1,\dagger}$, Xinyan Chen$^{2}$, Zhipeng Wang$^{1,2}$, Bin Cheng$^{1,2,*}$, and Bin He$^{1,2}$%
\thanks{$^{1}$College of Electronic and Information Engineering, Tongji University, Shanghai, China.}%
\thanks{$^{2}$Shanghai Research Institute for Intelligent Autonomous Systems, Tongji University, Shanghai, China.}%
}

\begin{document}

\maketitle
\thispagestyle{empty}
\pagestyle{empty}

\begin{abstract}
Complex multi-agent control tasks remain challenging for traditional rule-based and model-based approaches, motivating the adoption of learning-based methods.
However, learning-based methods often struggle with sim-to-real transfer because they rely on accurate dynamics modeling or system identification and learn policies in low-level control spaces that are highly sensitive to dynamics mismatch, making them costly and fragile in complex environments.
To address this issue, we propose a sim-to-real method for multi-agent control, which is insensitive to dynamics mismatch via effect alignment.
Our method combines random environmental structure with discrete semantic actions through closed-loop control, elevating policy learning to a semantic abstraction level. 
Additionally, we develop an action synchronization mechanism that mitigates inter-agent action timing mismatches, thereby enhancing the temporal consistency of the system.
Experiments on four multi-agent navigation tasks demonstrate that our method substantially improves training efficiency over mainstream transfer methods and achieves higher success rates in real-world scenarios, thereby improving the robustness and deployment stability of multi-agent systems under dynamics mismatch.
\end{abstract}
\section{Introduction}
In recent years, multi-agent reinforcement learning (MARL) has achieved substantial breakthroughs in domains such as complex games and distributed control, and has emerged as a principal method for addressing multi-agent swarm control problems~\cite{ekechi2025survey}. 
Benefiting from the strong representational capacity of deep neural networks, agents can exhibit outstanding performance on challenging tasks that are often beyond the reach of traditional control methods, including unmanned aerial vehicle swarms and multi-robot collaborative warehousing~\cite{zhao2024graph}.
Compared to traditional control methods, MARL demonstrates superior adaptability and robustness in handling high-dimensional state spaces and multi-robot cooperative interactions.
Therefore, deploying MARL in practical multi-robot systems is a promising avenue to achieve embodied intelligence.

Currently, MARL training is conducted predominantly in simulation, as simulation enables faster, safer, and more cost-effective training data collection than real-world deployment. 
However, the main challenge in deploying MARL in the real world is that policies are trained in simulation but executed under different real-world dynamics.
Sim-to-real transfer methods are required to address the issues caused by the discrepancies between the training and deployment environments in reinforcement learning.
The prevailing methods for mitigating the reality gap~\cite{da2025survey} include domain randomization~\cite{akkaya2019solving} and domain adaptation~\cite{chebotar2019closing}.
Domain randomization perturbs simulator dynamics parameters to improve policy robustness.
Domain adaptation, on the contrary, leverages real-world data or interactions to adapt policies or dynamics models to the target domain.
Despite these abilities to enhance the fidelity of simulation, the real world still exhibits characteristics that are difficult to model, such as nonlinear friction and hysteresis in control signals~\cite{gao2022review, WANG2024103166}.
Besides, the intrinsic mismatch and non-stationarity of models in MARL~\cite{zhu2024survey,shi2024sample} can make policies more vulnerable when confronted with physical disturbances in real environments.

To address the limitations of existing methods in sim-to-real transfer for MARL, 
rather than explicitly modeling or covering possible discrepancies between simulation and the real world,
we propose insensitive to dynamics mismatch via effect alignment (IDEA), a method designed to be robust to physical discrepancies between simulation and the real world. 
The key intuition is inspired by biological motor control, where robust locomotion is often preserved under environmental changes through internal adjustment rather than complete relearning.
Motivated by this observation, we seek to learn policies whose action effects remain consistent across diverse environments, rather than overfitting to specific simulated conditions.

\begin{figure*}[ht]
  \begin{center}
    \centerline{\includegraphics[width=\linewidth, keepaspectratio]{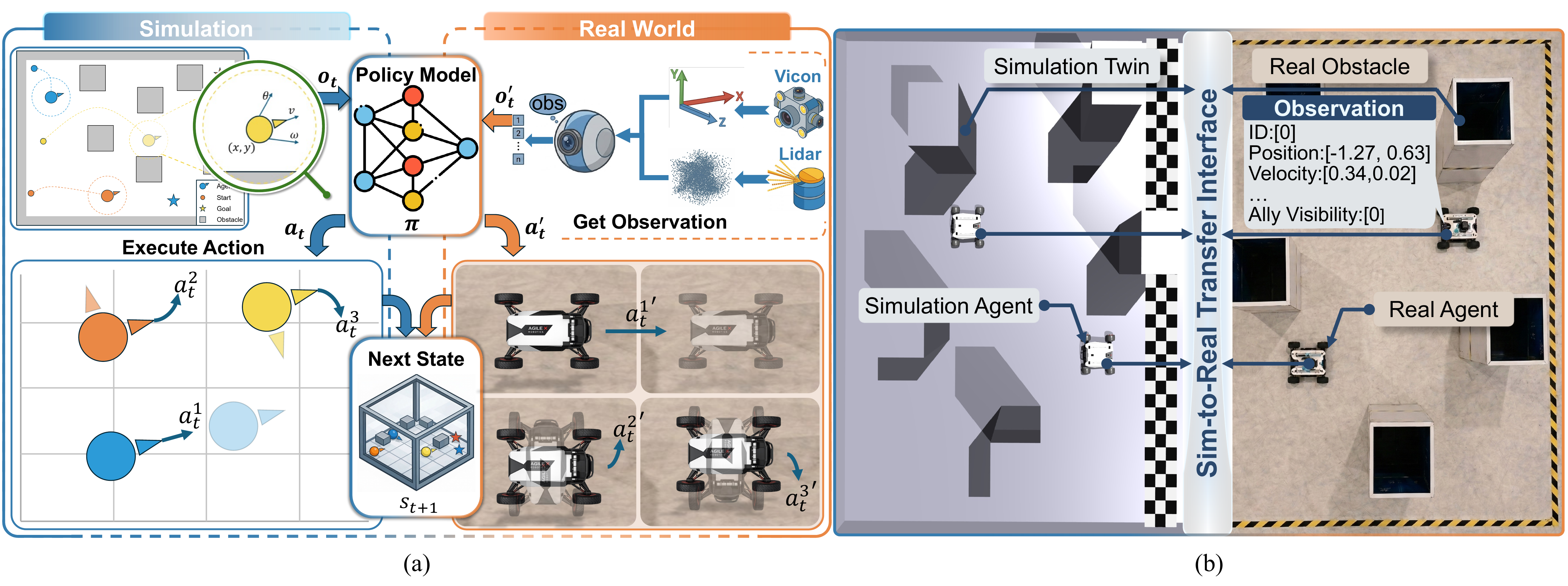}}
    \caption{
    Overview of the proposed sim-to-real transfer framework and experimental setup. 
    (a) The execution pipeline illustrating the zero-shot deployment of the simulation-trained policy to the real world physical agents. Here, $(x,y,\theta,v,\omega)$ indicate the agent's coordinates, yaw angle, linear velocity, and angular velocity, $o$ and $a$ are defined in Sec.~\ref{s_3_a}. (b) A comparison of the simulation digital twin and the corresponding physical environment, highlighting the observation and execution interface bridging the reality gap.
    }
    \label{Work Flow}
  \end{center}
\end{figure*}

Specifically, we train the policy across parallel simulated environments with diverse geometric structures to ensure that the policy exhibits strong adaptability across different structures. 
Furthermore, to avoid the effect of mismatched physical parameters, we combine discrete semantic actions with closed-loop control, thereby aligning the execution effects of actions in both the real world and simulation.
Additionally, we introduce a communication-based action synchronization mechanism among agents to synchronize the execution timing of actions across agents during real-world deployment.
To this end, we build a computationally efficient high-speed training platform based on the Isaac Gym~\cite{makoviychuk2021isaac} simulation engine to validate our method and provide diverse geometric structures in parallel environments.
We train all policies on this platform and transfer them to the real world. 
The overall transfer process is illustrated in Fig.~\ref{Work Flow}.
Our method is evaluated across four challenging multi-agent navigation scenarios, compared with the classic domain randomization and domain adaptation baselines. 
The experimental results demonstrate the advantages of our method in both training and deployment.
While achieving faster convergence and a higher success rate during training, our method improves the success rate of all tasks in the real world by more than 20\% over the baseline methods.
Furthermore, our method achieves zero collisions in the real world and zero errors in action execution timing among agents.

In summary, the main contributions of our work are as follows:
\begin{enumerate}
    \item We propose a novel sim-to-real transfer method that effectively reduces state-transition errors by aligning action effects between simulation and the real world.
    \item We introduce an inter-agent communication mechanism that synchronizes the execution timing of actions at each time step in actual deployment scenarios through communication among agents.
    \item We build a highly efficient parallel simulation training platform, which allows us to achieve large batch parallel training despite limited computing resources, thereby supporting our method.
\end{enumerate}

\section{Related Works}
\subsection{Applications of MARL in Control}
Classical control methods in real-world robotic tasks involving complex interactions are often unsatisfactory. 
Researchers have turned to reinforcement learning, which has achieved notable success in robot control, games, fine-tuning of models~\cite{hwangbo2019learning,kaufmann2023champion,ouyang2022training,vinyals2019grandmaster,zhang2025bridging}.
MARL extends this method to multi-agent systems and has been powerful in addressing complex coordination problems in robotics~\cite{gronauer2022multi}.
Representative algorithms such as multi-agent proximal policy optimization (MAPPO)~\cite{NEURIPS2022_9c1535a0}, which follows the centralized training and decentralized execution (CTDE) paradigm~\cite{kraemer2016multi}, leverage global state during training, while operating on local observations during execution.
Recent advances demonstrate their effectiveness in a variety of challenging cooperative tasks, including multi-agent path planning~\cite{sartoretti2019primal,damani2021primal}, coordinated unmanned ground vehicle (UGV) swarms~\cite{lin2024autonomous}, and cooperative locomotion~\cite{haarnoja2024learning}. 
However, directly training MARL policies within the native action spaces of agents can severely hinder training convergence.
To overcome this limitation, hierarchical reinforcement learning can abstract low-level control signals into high-level actions, allowing the policy to make decisions at a higher level~\cite{nachum2018data,liu2025slap}. 
Hierarchical methods have been used to accelerate exploration and solve long-term tasks in sparse reward environments~\cite{hu2025slac}.
Despite these achievements, transferring MARL policies to physical multi-agent systems remains severely constrained by the reality gap~\cite{dulac2021challenges}.
\subsection{Sim-to-Real Transfer}
To bridge the reality gap induced by discrepancies in physical parameters, domain randomization methods train policies across a distribution of simulated environments whose parameters are perturbed  periodically~\cite{tobin2017domain,muratore2019assessing}. 
A fundamental baseline in this paradigm is dynamic randomization (DR)~\cite{peng2018sim}, which encourages the emergence of robust behaviors in target environments with varying physical characteristics.
Subsequent methods have introduced adaptive sampling strategies that dynamically refine randomization distributions based on the experiences of the target environment or the performance of the agent~\cite{chebotar2019closing,mehta2020active}.

In contrast, rather than solely relying on randomization of parameters, recent methods emphasize online adaptation and system identification to handle dynamics mismatches. 
These methods typically leverage meta-learning frameworks~\cite{wang2016learning} or history-dependent modules to implicitly infer latent physical properties during deployment. 
For example, rapid motor adaptation (RMA) paradigms~\cite{qi2023hand,kumar2021rma} train policies to estimate the extrinsic impact of the environment from recent proprioceptive histories.
Different from purely implicit representations, other methods incorporate explicit hybrid internal models (HIM)~\cite{long2023hybrid}, using predicted simulated responses to guide agile actions in the real world.
This paradigm has been extended to learn exploratory motor skills to explicitly identify system dynamics on the fly, allowing rapid adaptation to novel deployment environments~\cite{margolis2023learning,memmel2024asid}.
Although these methods have reduced the reality gap, they still have shortcomings.
Domain randomization methods often lead to a decrease in convergence speed.
On the other hand, online adaptation methods heavily rely on complex recurrent architectures or extensive history encoders. 
These dependencies can introduce delay during inference and exacerbate compounding errors in long-horizon tasks.
\section{Method}
In this section, we present the architecture and theoretical foundations of IDEA, a novel method for multi-agent control designed for robust sim-to-real transfer in MARL.
The core philosophy of our method is to abstract away the need for highly accurate modeling of complex low-level physical dynamics.
Instead, we enable policies to operate at a high semantic level, where discrete semantic actions induce aligned state transitions in both simulation and the real world.
By aligning these discrete action effects, our method mitigates the reality gap and facilitates zero-shot transfer for decentralized multi-agent systems.

\begin{figure*}[ht]
  \begin{center}
    \centerline{\includegraphics[width=\textwidth,height=\textheight,keepaspectratio]{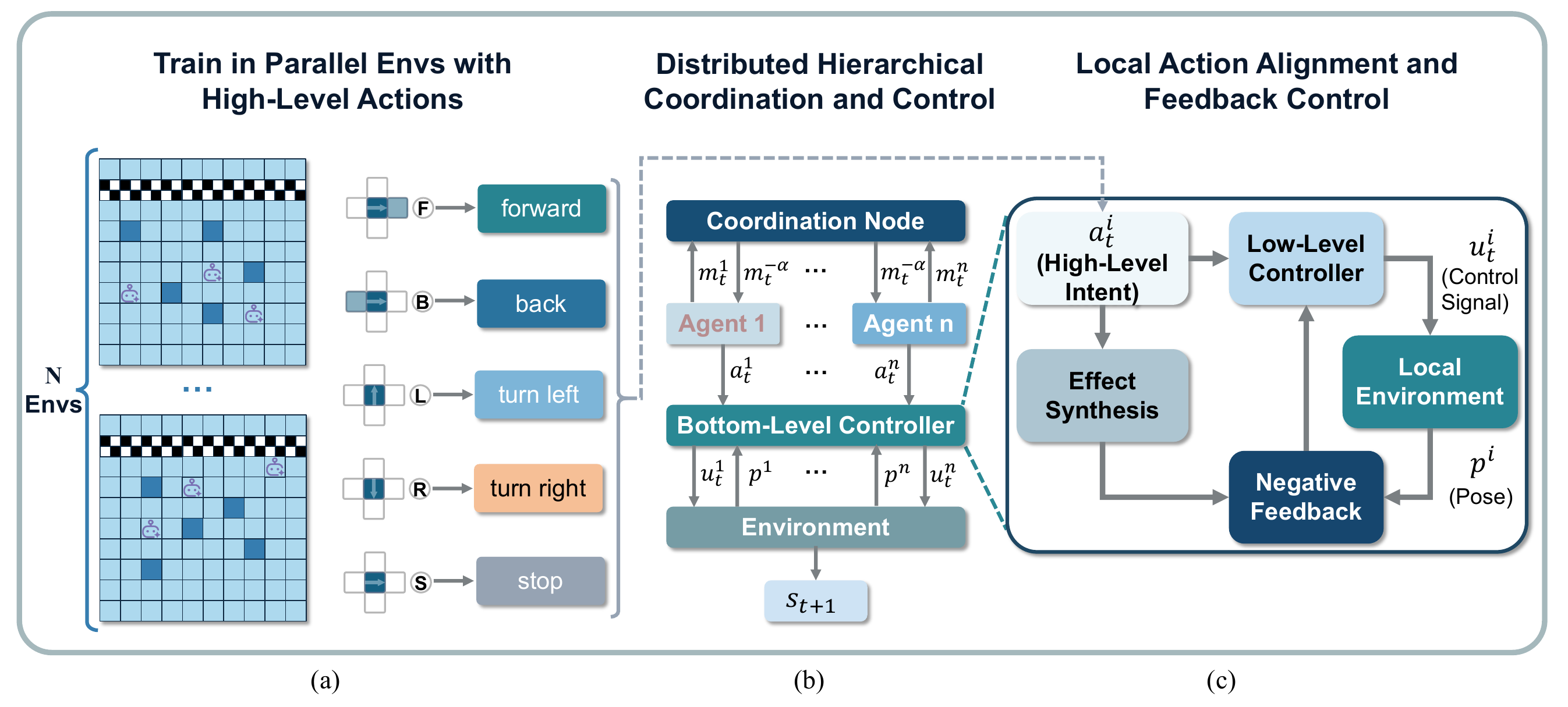}}
    \caption{ 
    The architecture of IDEA.
    (a) The simulation training phase, utilizing parallel environments and discretized high-level semantic actions. 
    (b) The distributed hierarchical control structure deployed on the multi-agent system, translating coordinated intents into bottom-level control signals. Here, $m$ and $\alpha$ are defined in~\ref{s_3_c}.
    (c) The local action alignment module, employing a closed-loop negative feedback mechanism to ensure high-fidelity physical execution of the intended semantic actions.
    }
    \label{Framework}
  \end{center}
\end{figure*}

To achieve this, the proposed method integrates three critical mechanisms. 
First, effect alignment serves as the core theoretical bridge of our method. By mapping discrete semantic actions to continuous control through low-level controllers, we bound the sim-to-real performance gap without explicitly modeling low-level physical parameters.
Second, to ensure that the theoretical assumptions of discrete time MARL hold during physical deployment, we introduce a strict action synchronization mechanism that eliminates temporal inconsistencies caused by decentralized execution. 
Third, we employ parallel environments with different geometric structures during simulation to prevent policies from overfitting to specific layouts, allowing them to generalize to diverse real-world environments without fine-grained parameter tuning.
The overall architecture is illustrated in Fig.~\ref{Framework}.

\subsection{Problem Formulation}
\label{s_3_a}
This paper models the multi-agent coordination problem across diverse geometric configurations as a context-conditioned Decentralized Partially Observable Markov Decision Process (Dec-POMDP).
This formulation extends the standard Dec-POMDP~\cite{bernstein2002complexity} by explicitly conditioning the environment dynamics on a static geometric context, defined by the tuple 
$\mathcal{M} = 
\langle 
\mathcal{N}, \mathcal{S}, \mathcal{A}, \mathcal{P}, \mathcal{R}, \mathcal{O}, \mathcal{C}, \gamma 
\rangle$.
Here, $\mathcal{N} = \{1, \ldots, N\}$ denotes the finite set of agents.
To model domain diversity for sim-to-real transfer, we introduce a random context variable $c \in \mathcal{C}$ sampled from a training distribution $\mathcal{D}_{\text{train}}$.
Each context $c$ encapsulates the geometric configuration of the environment, including initial agent poses and terrain layouts, which are discretized into a Boolean tensor $H \in \mathbb{B}^{l \times w}$. 
For each element $H_{j,k}$, a value of $1$ indicates the presence of an obstacle in that grid cell, while $0$ denotes free space. 
A geometric configuration $c$ is sampled at the beginning of each episode and remains strictly fixed throughout the execution.

At time step $t$, based on the global state $s_t \in \mathcal{S}$, each agent $i$ receives a local observation $o_t^i \in \mathcal{O}^i$ and selects a discrete semantic action $a_t^i \in \mathcal{A}^i$ according to its decentralized policy $\pi^i(a_t^i \mid o_t^i)$. 
Executing the joint action $\boldsymbol{a}_t = \langle a_t^1, \dots, a_t^N \rangle$ triggers a transition to the next state $s_{t+1} \sim \mathcal{P}(\cdot \mid s_t, \boldsymbol{a}_t, c)$ and produces a shared global reward $r_t = \mathcal{R}(s_t, \boldsymbol{a}_t, c)$. 
Following the CTDE paradigm, the overall objective is to learn an optimal joint policy $\boldsymbol{\pi} = \langle \pi^1, \dots, \pi^N \rangle$ that maximizes the expected discounted return across the entire distribution of geometric configurations:
\begin{equation}
\label{eq_1}
\max_{\boldsymbol{\pi}}\; J(\boldsymbol{\pi})= 
\mathbb{E}_{c \sim \mathcal{D}_{\mathrm{train}}}
\left[ V^{\boldsymbol{\pi}}(c) \right],
\end{equation}
\begin{equation}
\label{eq_2}
V^{\boldsymbol{\pi}}(c) = 
\mathbb{E}_{\tau \sim (\mathcal{M}(c),\,\boldsymbol{\pi})}
\!\left[
\sum_{t=0}^{\infty} {\gamma}^t \mathcal{R}(s_t, \boldsymbol{a}_t, c)
\right]. 
\end{equation}
\subsection{Effect Alignment}
\label{s_3_b}
To enable sim-to-real transfer without explicitly modeling physical parameters, we adopt diverse geometric structure combined with discrete semantic actions, allowing the policy to operate at a high semantic level. 
The joint policy outputs a joint discrete semantic action $\boldsymbol{a}_t \in \mathcal{A}$, sampled as
$\boldsymbol{a}_t \sim \boldsymbol{\pi} (\boldsymbol{a} \mid \boldsymbol{o}_t)$.
In real-world deployment, a decentralized low-level controller $k$ maps the semantic action to continuous control commands, resulting in
$\boldsymbol{u}_t \in \mathcal{U}$
via $\boldsymbol{u}_t = k(\boldsymbol{o}_t, \boldsymbol{a}_t)$. 
As a result, the environment transition kernel is transformed into a closed-loop induced transition kernel by Eq.~\eqref{eq_3}.
\begin{equation}
\label{eq_3}
\tilde{\mathcal{P}}(s_{t+1} \mid s_t,\boldsymbol{a}_t,c)
=\mathbb{E}_{\boldsymbol{o}_t \sim \mathcal{O}} 
\left[
\mathcal{P}\bigl(s_{t+1} \mid s_t, \boldsymbol{u}_t,\, c \bigr) 
\right].
\end{equation}

During training, the policy $\boldsymbol{\pi} (\boldsymbol{a} \mid \boldsymbol{o})$ is optimized under the induced transition kernel $\tilde{\mathcal{P}}_{\text{sim}}$,
whereas in real-world deployment it operates under $\tilde{\mathcal{P}}_{\text{real}}$. 

Consequently, the main challenge of sim-to-real transfer is to minimize the discrepancy between $\tilde{\mathcal{P}}_{\text{sim}}$ and $\tilde{\mathcal{P}}_{\text{real}}$. 
For continuous state spaces, we assume the induced transition mismatch is bounded by $\epsilon$:
\begin{equation}
\label{eq_4}
W_1(\tilde{\mathcal{P}}_{\text{sim}}(\cdot \mid s,\boldsymbol{a}, c), \tilde{\mathcal{P}}_{\text{real}}(\cdot \mid s,\boldsymbol{a}, c)) \le \epsilon.
\end{equation}

By the definition of the Kantorovich formulation of optimal transport~\cite{peyre2019computational}, the $W_1$ distance is the upper bound of the expected distance between any randomly coupled random variables.
Assuming the simulated and real states are aligned at time step $t$, the execution of the same semantic action $\boldsymbol{a}_t$ yields the next states.
If agent-level closed-loop control bounds the expected metric distance between these next states:
\begin{equation}
\label{eq_5}
\mathbb{E} \left[
\left\lVert s_{\text{sim}}^{t+1} - s_{\text{real}}^{t+1} \right\rVert_{1} \right] \le \epsilon,
\end{equation}
then the Eq.~\eqref{eq_4} follows from Eq.~\eqref{eq_5}.
Furthermore, assuming identical reward functions across domains and the value function $V_\text{sim}^\pi$ is $L_V$-Lipschitz continuous with respect to the state metric, and applying the Simulation Lemma for continuous state spaces~\cite{asadi2018lipschitz}, the performance gap under simulation and real world dynamics can be bounded by:
\begin{equation}
\label{eq_6}
\left\lvert V_{\text{real}}^\pi(c) - V_{\text{sim}}^\pi(c) \right\rvert
\le  \cfrac{\gamma L_V}{1 - \gamma} \epsilon.
\end{equation}

This theoretical result implies that as long as the closed-loop controller onboard agents effectively aligns the execution effects of actions, which     bounds the mismatch term $\epsilon$ in Eq.~\eqref{eq_5}, the policy can retain strong performance in real-world deployment, significantly mitigating the reality gap.

\subsection{Action Synchronization}
\label{s_3_c}
Our theoretical performance bound in Section~\ref{s_3_b} is critically based on the assumption that the joint semantic action $\boldsymbol{a}_t$ is executed simultaneously, mirroring the strictly discrete time steps of the simulated Dec-POMDP. 
However, when deploying MARL policies in real world settings with decentralized execution, the system inevitably suffers from action asynchrony caused by controller clock drift, sensor latency, heterogeneous on-board computation, and communication delays.
These factors can lead to significant temporal discrepancies across agents at the same nominal time step, violating the state-action alignment required for Eq.~\eqref{eq_5} and severely degrading coordinated behavior.

To bridge this gap and physically enforce the joint action assumption, we introduce a synchronization mechanism that guarantees step-wise action alignment among agents during execution.
All agents are deployed within the same local wireless network and communicate via peer-to-peer messaging.
Let the set of agents be denoted by $\mathcal{N} = \{1, 2, ..., N\}$. An agent $\alpha \in \mathcal{N}$ is designated as a coordination node, responsible for determining the exact execution timing. 
Before executing its individual semantic action $a_t^i$ at time step $t$, each agent $i \neq \alpha$ sends a readiness signal $m_t^i$ to the coordination node.
Once agent $\alpha$ confirms that all agents are ready, it broadcasts an execution signal to all other agents, upon which synchronized joint action execution proceeds simultaneously.
\subsection{Training in Parallel Environment}
\label{s_3_d}
To support our method and accelerate training under limited computational resources, we extend the Isaac Gym simulator by customizing its interaction logic and tensor-based data pipeline.
This design preserves the high-throughput performance of the simulator while enabling parallel multi-agent simulation across diverse environments.
Our implementation treats the simulator as a generator of diverse geometric structures that support agent-environment interaction.
At the start of training, different geometric configurations, represented by the context variable $c \sim \mathcal{D}_{\text{train}}$, are assigned to all parallel environments, and all the data needed during training are processed using tensors that reside entirely on the GPU.

Furthermore, to preserve the validity of the context-conditioned Dec-POMDP, we adopt a strict episode-level reset scheme.
Specifically, the geometric configuration $c$ for each parallel environment is sampled only at instantiation and remains static across episode resets.
To ensure sufficient environmental diversity for the training distribution $\mathcal{D}_{\text{train}}$, we scale up the total number of parallel environments rather than resampling configurations per episode. 
This design effectively improves policy generalization while mitigating the computational overhead and training instability typically associated with high-frequency, per-episode randomization.
\section{Experiments}
\subsection{Experimental Setup}
\begin{figure}[ht]
  \begin{center}
    \centerline{\includegraphics[width=\linewidth,keepaspectratio]{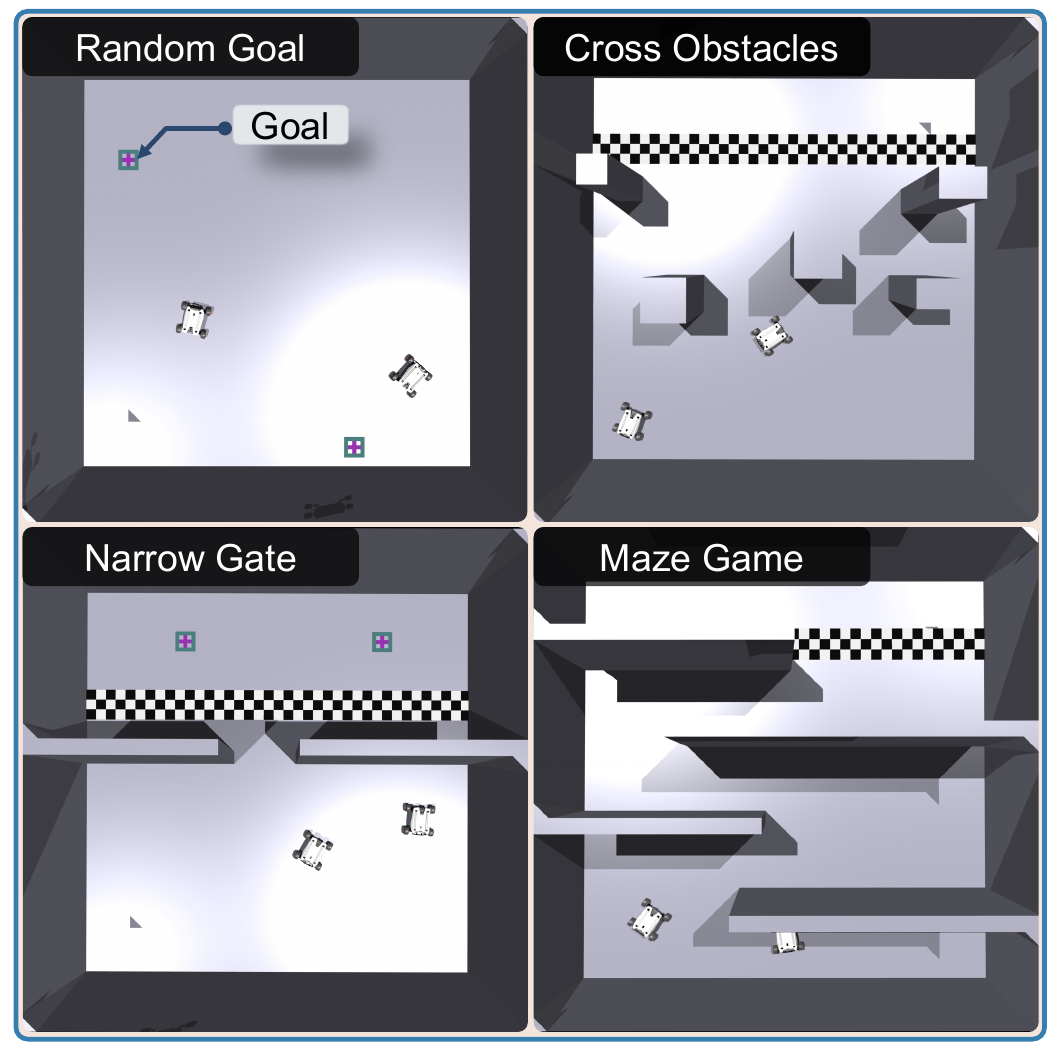}}
    \caption{Overview of the evaluation multi-agent navigation scenarios with increasing geometric and coordination complexity.}
    \label{tasks}
  \end{center}
\end{figure}

We use a two-UGV system for real-world experiments, where each vehicle has four degrees of freedom.
We use MAPPO~\cite{NEURIPS2022_9c1535a0} as the base MARL algorithm.
We construct four tasks, as illustrated in Fig.~\ref{tasks}.
The local observation used for policy learning comprises the agent’s ego-state and its surrounding environmental context.
During simulation, these states are directly accessed via the simulator's application programming interface.
For real-world transfer, we capture the agent's ego-state using a Vicon motion capture system, while the local environment is perceived by an onboard 2D lidar.
To reduce the observation discrepancy between simulation and the real world, the raw point cloud generated by the lidar is processed in real time into an occupancy grid map, closely approximating the map used in simulation.
Furthermore, real-world episodes use the same control horizon as in training, with the policy outputting action commands at 5 Hz.
Additional details of the experimental setup and hyperparameters are provided in Appendix A.
\subsection{Task Definition}
In all evaluation scenarios, as shown in Fig.~\ref{tasks}, the fundamental objective of agents is navigating to their respective destination while maintaining strictly collision-free trajectories. 
The tasks are designed with increasing geometric and coordination complexity:
(1) \textit{Random Goal.}\quad A workspace without obstacles where agents and their target positions are uniformly randomized, serving as a baseline for basic decentralized navigation.
(2) \textit{Cross Obstacles.}\quad Agents spawn on one side of the environment and must traverse a dense cluster of static obstacles with randomized dimensions and poses to reach the opposite side.
(3) \textit{Narrow Gate.}\quad The workspace is segmented by a partition with a narrow gate that allows only a single passage. 
Agents must learn to yield and coordinate sequentially to pass the gate.
(4) \textit{Maze Game.}\quad A highly constrained maze composed of randomly placed narrow walls. Agents must perform long-horizon spatial reasoning and coordinated pathfinding to reach the far side.

\subsection{Simulation Training}
\begin{figure}[ht]
  \begin{center}
    \centerline{\includegraphics[width=\linewidth,keepaspectratio]{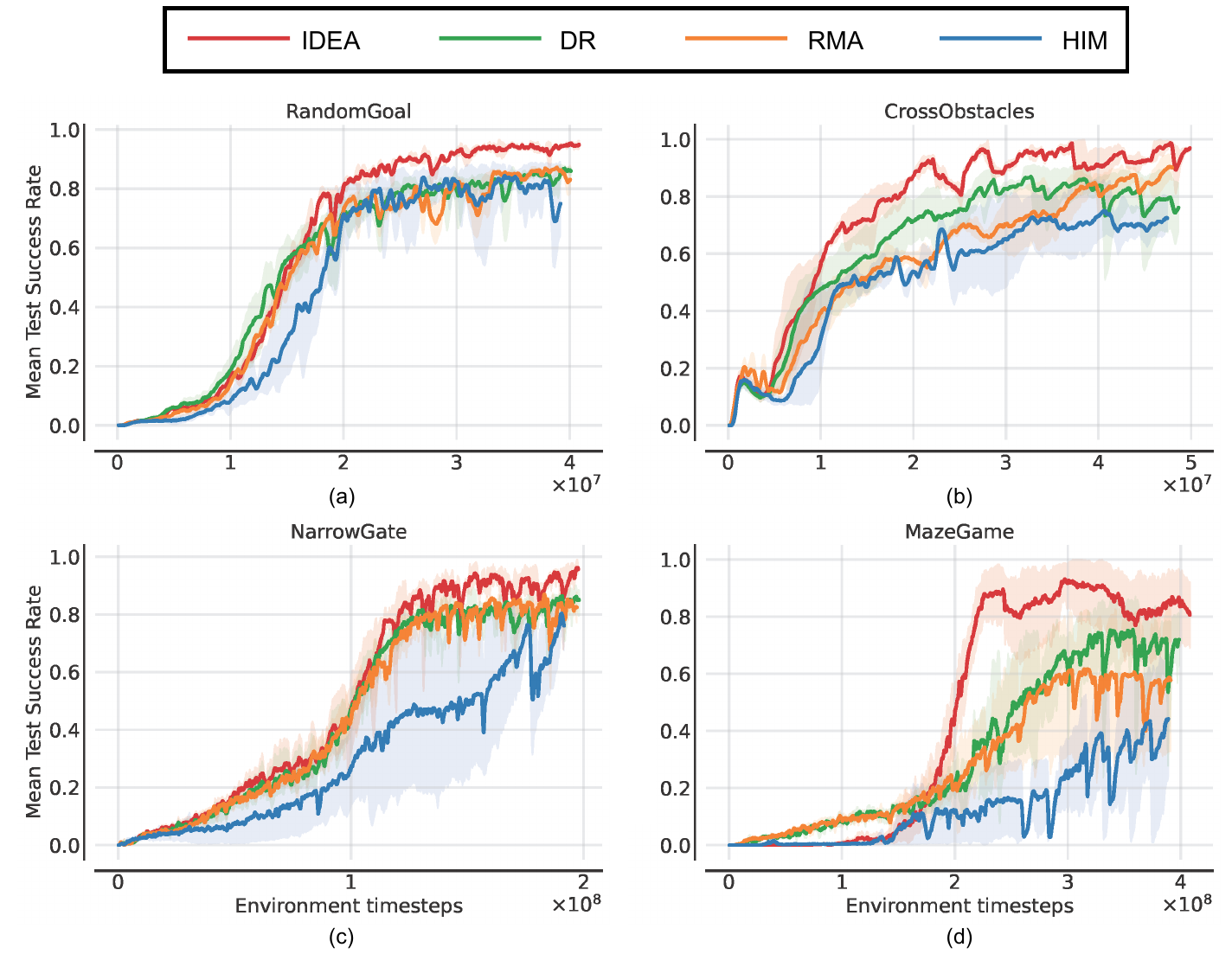}}
    \caption{Learning curves across four simulated navigation tasks. For fair evaluation, all experiments are conducted with three random seeds, and results are reported as means with 95\% confidence intervals}
    \label{train}
  \end{center}
\end{figure}

We evaluate policies across all tasks, benchmarking them against representative sim-to-real baselines.
In particular, we consider two categories of methods: domain randomization methods and domain adaptive methods. 
Specifically, we compare against three baselines: DR~\cite{peng2018sim}, RMA~\cite{qi2023hand,kumar2021rma}, and HIM~\cite{long2023hybrid}.
For each experiment, performance is averaged over multiple models trained with varying random seeds.
As shown in Fig.~\ref{train}, our method outperforms the baselines across the evaluated tasks.
Compared to methods that rely on standard domain randomization or adaptation, our method achieves substantially faster convergence while maintaining comparable or superior final success rates.
\subsection{Zero-Shot Transfer}

\begin{figure}[htbp]
  \begin{center}
    \centerline{\includegraphics[width=\linewidth,keepaspectratio]{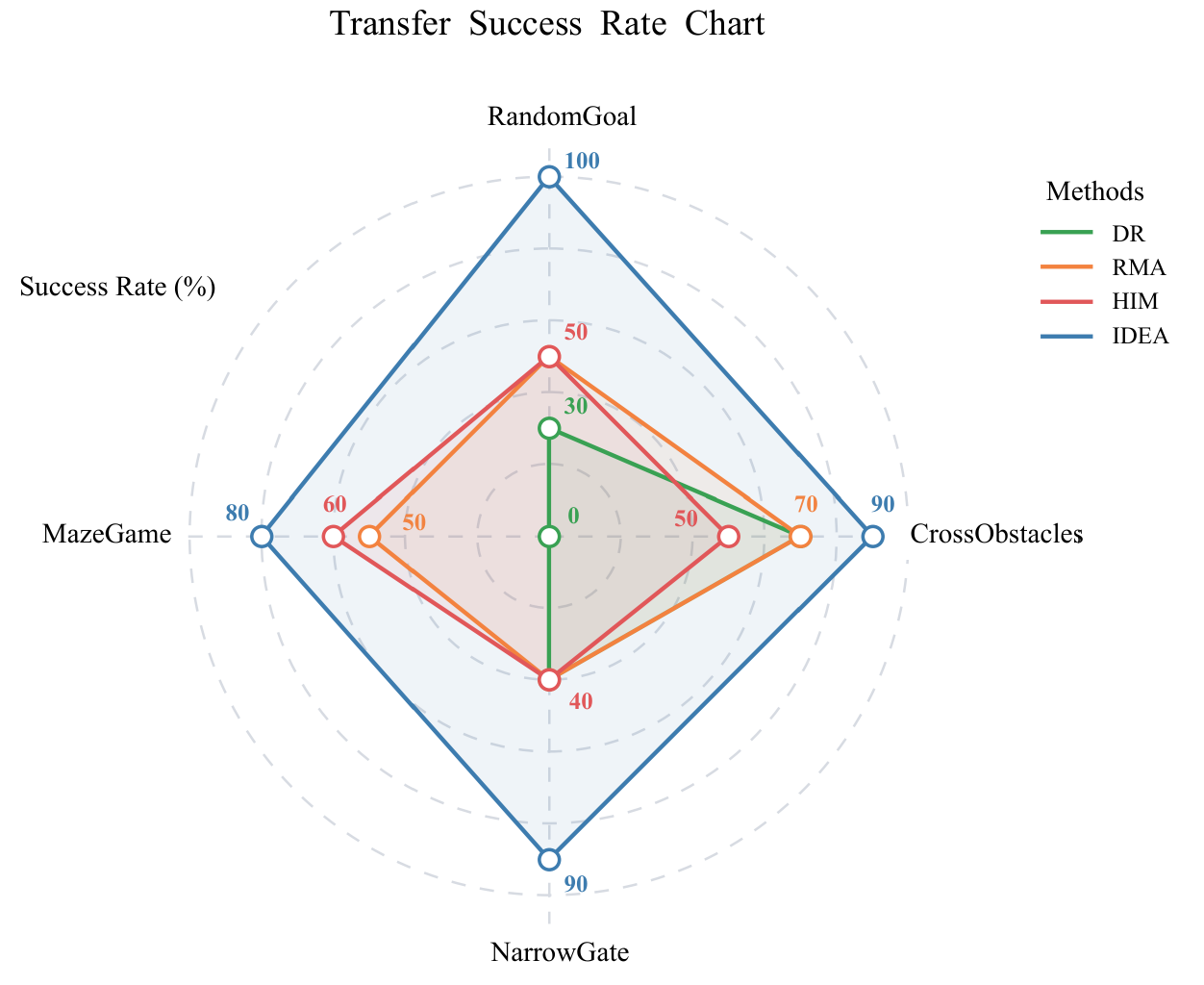}}
    \caption{Radar chart illustrating the zero-shot transfer success rates (\%) of the evaluated methods in the real world. Our proposed method demonstrates superior transferability across all tasks compared to the baselines.}
    \label{transfer}
  \end{center}
\end{figure}

\begin{table*}[ht]
\centering
\caption{Action effects in all methods and simulation.}
\label{tab:sim_real_action_compare}
\begin{threeparttable}
\setlength{\tabcolsep}{8pt}
\renewcommand{\arraystretch}{1.2}
\begin{tabular}{ccccccc}
\toprule
\textbf{Action} & \textbf{Unit} & \textbf{Simulation} & \textbf{IDEA} & \textbf{DR} & \textbf{RMA} & \textbf{HIM} \\
\midrule
Ahead(full step) & m & 0.20 & \textbf{0.194 $\pm$ 0.002} & 0.134 $\pm$ 0.013 & 0.140 $\pm$ 0.015 & 0.145 $\pm$ 0.028 \\
Ahead(half step) & m & 0.10 & \textbf{0.096 $\pm$ 0.002} & 0.068 $\pm$ 0.009 & 0.066 $\pm$ 0.015 & 0.069 $\pm$ 0.011 \\
Back(full step) & m & -0.20 & \textbf{-0.195 $\pm$ 0.003} & -0.133 $\pm$ 0.014 & -0.136 $\pm$ 0.019 & -0.128 $\pm$ 0.014 \\
Back(half step) & m & -0.10 & \textbf{-0.095 $\pm$ 0.002} & -0.070 $\pm$ 0.011 & -0.068 $\pm$ 0.013 & -0.071 $\pm$ 0.013 \\
Left & rad & 0.15 & \textbf{0.147 $\pm$ 0.002} & 0.135 $\pm$ 0.009 & 0.127 $\pm$ 0.064 & 0.113 $\pm$ 0.014 \\
Right & rad & -0.15 & \textbf{-0.146 $\pm$ 0.001} & -0.134 $\pm$ 0.015 & -0.130 $\pm$ 0.016 & -0.124 $\pm$ 0.044 \\
\bottomrule
\end{tabular}
\begin{tablenotes}
\footnotesize
\item
\end{tablenotes}
\end{threeparttable}
\end{table*}
To evaluate the real world transferability of the learned policies, we conduct zero-shot deployment experiments where models trained exclusively in simulation are directly deployed on real platforms without any fine-tuning.
To reduce the effect of random variation, we collect 10 distinct trajectories for each method in all tasks.
As depicted in Fig.~\ref{transfer}, our proposed method consistently achieves the highest transfer success rates in all evaluated scenarios.

In contrast, the baseline methods exhibit significant performance degradation. 
Notably, DR yields the lowest success rates. 
A possible reason is that DR is fundamentally constrained by the parametric capabilities of the simulator. 
It can only randomize predefined physical variables, leaving the policy completely vulnerable to structural unmodeled dynamics.
RMA and HIM performed similarly because both depend on local history for adaptation.
This structural design inevitably introduces an adaptation lag, meaning that agents can only adjust their policies after an unmodeled disturbance has already affected their dynamics.
To further substantiate this claim, we analyze transfer behaviors across three dimensions: action effects consistency, action temporal differences between agents, and types of task failure.

First, we quantify the fidelity of physical execution by comparing the real world action effects against their expected results in simulation.
The comparison is shown in Table~\ref{tab:sim_real_action_compare}. 
Although baselines suffer from severe execution mismatches, our method substantially reduces the discrepancy between simulated action effects and their physical execution, particularly for translational motions.
In real-world environments, the action mismatch is less than 5\%.

Second, we evaluate inter-agent temporal discrepancies by measuring the mean differences of the steps required to complete the tasks, shown in Table~\ref{tab:avg_step_gap}.
Our proposed method maintains zero difference across all environments, whereas the baseline methods exhibited a temporal discrepancy of up to 11.5 steps. 
This result indicates that our method not only achieves a higher success rate, but also preserves coordinated behavior after transfer.
\begin{table}[!htbp]
\centering
\caption{Average step difference of all methods.}
\label{tab:avg_step_gap}
\footnotesize
\setlength{\tabcolsep}{2.3pt}
\renewcommand{\arraystretch}{1.2}
\begin{tabular}{cccccc}
\toprule
\textbf{Tasks} & \textbf{Unit} & \textbf{DR} & \textbf{RMA} & \textbf{HIM} & \textbf{IDEA} \\
\midrule
RandomGoal & step & 11.25 $\pm$ 6.18 & 9.00 $\pm$ 2.16 & 7.83 $\pm$ 5.12 & \textbf{0} \\
CrossObstacles & step & 8.00 $\pm$ 3.90 & 6.30 $\pm$ 2.35 & 8.17 $\pm$ 4.07 & \textbf{0} \\
NarrowGate & step & 7.50 $\pm$ 3.42 & 7.00 $\pm$ 4.53 & 6.75 $\pm$ 3.20 & \textbf{0} \\
MazeGame & step & 5.60 $\pm$ 4.51 & 9.50 $\pm$ 5.26 & 11.50 $\pm$ 11.03 & \textbf{0} \\
\bottomrule
\end{tabular}
\end{table}

Finally, an analysis of failure modes in Table~\ref{tab:real_deploy_collision_timeout} reveals that our method achieves a strict 0\% collision rate while capping the timeout rate at a maximum of 20\% across all tasks.
In contrast, baselines suffer from substantially higher collision and timeout frequencies. 

In summary, transfer experiments demonstrate that by aligning action effects rather than relying on dynamics randomization during training or latent dynamics adaptation, our method substantially improves policy robustness, trajectory consistency, and safety during real-world deployment.
\begin{table}[htbp]
\centering
\scriptsize
\caption{Collision rate and timeout rate of all methods.}
\label{tab:real_deploy_collision_timeout}
\setlength{\tabcolsep}{4pt}
\renewcommand{\arraystretch}{1.15}
\begin{tabular}{c c cccc}
\toprule
\textbf{Method} & \textbf{Type} 
& \shortstack{\textbf{Random}\\\textbf{Goal}}
& \shortstack{\textbf{Cross}\\\textbf{Obstacles}}
& \shortstack{\textbf{Narrow}\\\textbf{Gate}}
& \shortstack{\textbf{Maze}\\\textbf{Game}} \\
\midrule
\multirow{2}{*}{DR}
& Collision & 20\% & 20\% & 0\%  & 50\% \\
& Timeout & 50\% & 10\% & 60\% & 50\% \\
\midrule
\multirow{2}{*}{RMA}
& Collision & 0\%  & 20\% & 10\% & 30\% \\
& Timeout & 50\% & 10\% & 50\% & 20\% \\
\midrule
\multirow{2}{*}{HIM}
& Collision & 10\% & 20\% & 0\%  & 40\% \\
& Timeout & 40\% & 30\% & 60\% & 0\%  \\
\midrule
\multirow{2}{*}{IDEA}
& Collision & \textbf{0\%}  & \textbf{0\%}  & \textbf{0\%}  & \textbf{0\%}  \\
& Timeout & \textbf{0\%}  & \textbf{10\%} & \textbf{10\%} & \textbf{20\%} \\
\bottomrule
\end{tabular}
\end{table}
\section{Conclusions}
In this paper, we introduced IDEA, a method for zero-shot sim-to-real transfer in multi-agent control. 
Extensive experiments across both simulated and real multi-agent platforms demonstrate that our method achieves high success rates and robust coordinated behavior, validating the effectiveness of focusing on high-level policies.
Despite these promising results, our method still has certain limitations. 
The proposed method relies on predefined discrete semantic action spaces and assumes that the geometric structure remains static within each episode. 
Additionally, during the experiment, we found that the action synchronization mechanism among agents introduces a temporal overhead, leading to increased overall task execution times.
Future work will focus on extending the principle of IDEA to continuous semantic action spaces for finer-grained control, optimizing the action synchronization protocol to mitigate temporal delays, and integrating dynamic obstacle tracking to accommodate highly non-stationary and interactive real-world environments.


\section*{Appendix}
\subsection{Experimental details}
\label{appendix_a}
The episode length ranges from 100 to 200 action steps, depending on task difficulty.
We design a unified reward formulation that guides multi-agent systems across diverse randomized terrains without task-specific reward engineering. 
The total reward of each agent consists of a distance-difference term $r_{\text{diff}}=d_{t-1} - d_t$, where $d$ denotes the distance from an agent to its goal or to the other agent, individual completion reward $r_{\text{pass}}$, team success reward $r_{\text{success}}$, time penalty $r_{\text{time}}$, a penalty for getting too close to the other agent $r_{\text{close}}=d_i - d_j$, and failure penalty $r_{\text{fail}}$, as defined in Eq.~\eqref{app:eq_1}:
\begin{equation}
\label{app:eq_1}
r_{\text{total}} = w_1r_{\text{diff}} + w_2r_{\text{close}} + r_{\text{time}} + r_{\text{pass}} + r_{\text{success}} + r_{\text{fail}}.
\end{equation}

The detailed configurations of the simulation environment, including the coefficients of the reward function and training hyperparameters, are summarized in Table~\ref{tab:reward_value}.
\begin{table}[ht]
\centering
\caption{Parameters in simulation experiments.}
\label{tab:reward_value}
\footnotesize
\setlength{\tabcolsep}{4pt}
\renewcommand{\arraystretch}{1.0}
\begin{tabular}{ccccc}
\toprule
\textbf{Reward} & \textbf{Coefficient} & \textbf{Value} & \textbf{Setting} & \textbf{Value}\\
\midrule
$ r_{\text{diff}} $ & 2.0  & $ d_t - d_{t-1} $ & map resolution & 0.25 m\\
$ r_{\text{close}} $ & -1.0 & $ d_{i,j} $ & simulation frequency & 60 Hz\\
$ r_{\text{time}} $ & 1.0  & $ -0.02 $ & hidden dim & 128\\
$ r_{\text{pass}} $ & 1.0  & $ 2.0 $ & batch size & 1024\\
$ r_{\text{success}} $ & 1.0  & $ 5.0 $ & entropy coef & 0.001\\
$ r_{\text{fail}} $ & 1.0  & $ -5.0 $ & learning rate & 0.0005\\
\bottomrule
\end{tabular}
\end{table}

\begin{figure}[ht]
  \begin{center}
    \centerline{\includegraphics[width=0.8\linewidth,keepaspectratio]{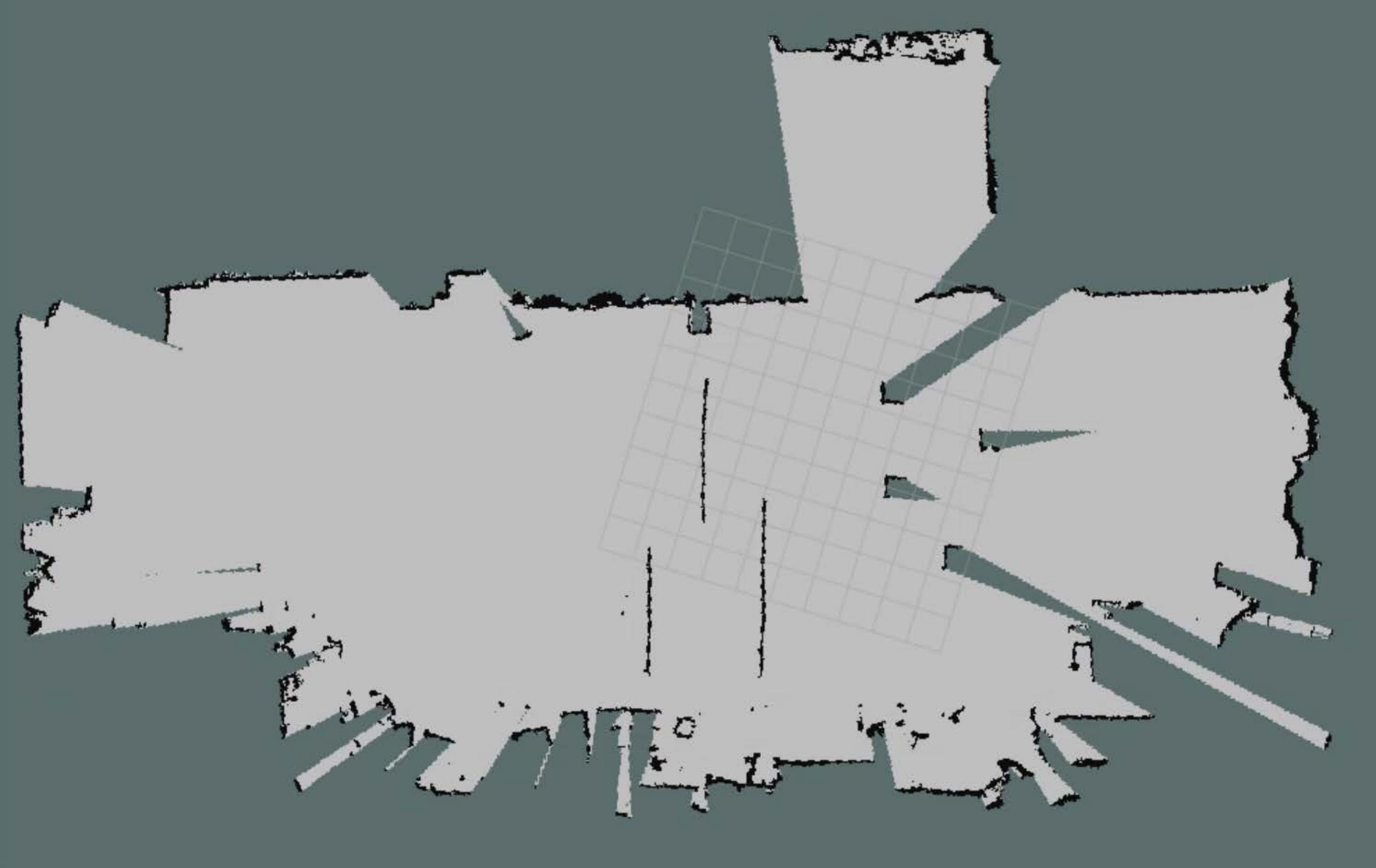}}
    \caption{An example of an actual scene's occupied grid map, taken from a MazeGame scene. The frame is taken from the baselink of one of the agents.}
    \label{occupancy map}
  \end{center}
\end{figure}

In the real world, to ensure the accuracy of the map, the lidar scans at a frequency of 20 Hz.
We first generate an occupied grid map with a resolution of 0.05 m, as shown in Fig.~\ref{occupancy map}, based on the lidar point cloud, and then downsample it to 0.25 m as the source of partial observation information.
\subsection{Theoretical Analysis of Effect Alignment}
In this section, we provide the formal proof that bounding the single-step expected state discrepancy via the closed-loop controller effectively bounds the infinite-horizon value function gap between simulation and the real world.

\textbf{Assumption.}\quad We assume that executing the same joint semantic action $\boldsymbol{a}$ from the same perfectly aligned state $s$ yields next states whose expected metric distance is bounded by a small constant $\epsilon$.
Besides, the state value function in simulation $V_{\text{sim}}^\pi(s, c)$ is $L_V$-Lipschitz continuous with respect to the state metric space:
\begin{equation}
     |V_{\text{sim}}^\pi(s_1, c) - V_{\text{sim}}^\pi(s_2, c)| \leq L_V \|s_1 - s_2\| \quad \forall s_1, s_2 \in \mathcal{S}.
\end{equation}

For simplicity of analysis, we assume the reward function $r(s, \boldsymbol{a}, c)$ is identical across both domains and is independent of the underlying low-level transition dynamics mismatch.

\textbf{Proof.}\quad We first relate the bounded expected state distance to the 1-Wasserstein distance ($W_1$) between the induced transition kernels. 
By its primal definition, the $W_1$ distance is the infimum of the expected distance over all possible joint distributions $\Gamma(P=\tilde{P}_{\text{sim}}, Q=\tilde{P}_{\text{real}})$ whose marginals are $P$ and $Q$:
\begin{equation}
    W_1(\tilde{P}_{\text{sim}}, \tilde{P}_{\text{real}}) = \inf_{\mu \in \Gamma} \mathbb{E}_{(s_{\text{sim}}', s_{\text{real}}') \sim \mu} \left[ \| s_{\text{sim}}' - s_{\text{real}}' \| \right].
\end{equation}

Since the parallel execution of the same semantic action $\boldsymbol{a}$ in both domains constitutes a valid specific coupling, the $W_1$ distance is naturally upper-bounded by the expected distance under this coupling:
\begin{equation}
    W_1(\tilde{P}_{\text{sim}}, \tilde{P}_{\text{real}}) \leq \mathbb{E}[\|s_{\text{sim}}' - s_{\text{real}}'\|] \leq \epsilon.
\end{equation}

Next, we evaluate the single-step expected value difference. 
By the Kantorovich-Rubinstein duality, the $W_1$ distance can be equivalently expressed as the supremum over all 1-Lipschitz functions:
\begin{equation}
    W_1(P, Q) = \sup_{\|f\|_L \leq 1} \left| \mathbb{E}_{x \sim P}[f(x)] - \mathbb{E}_{y \sim Q}[f(y)] \right|.
\end{equation}

Given that $V_{\text{sim}}^\pi$ is $L_V$-Lipschitz, the normalized function $V_{\text{sim}}^\pi / L_V$ is 1-Lipschitz. 
Applying the duality yields the single-step value error bound:
\begin{equation}
\begin{aligned}
&\left| \mathbb{E}_{s' \sim \tilde{P}_{\text{real}}}[V_{\text{sim}}^\pi(s')]
- \mathbb{E}_{s' \sim \tilde{P}_{\text{sim}}}[V_{\text{sim}}^\pi(s')] \right| \\
&\leq L_V \cdot W_1(\tilde{P}_{\text{sim}}, \tilde{P}_{\text{real}}) \leq L_V \epsilon.
\end{aligned}
\end{equation}

Finally, we roll out this single-step error over the infinite horizon using the Bellman expectation equation.
Let the maximum value discrepancy be $\Delta V = \Delta V(s) = |V_{\text{real}}^\pi(s) - V_{\text{sim}}^\pi(s)$. 
Evaluating the absolute difference:
\begin{equation}
    \Delta V(s) = \gamma \left| \mathbb{E}_{s' \sim \tilde{P}_{\text{real}}} [V_{\text{real}}^\pi(s')] - \mathbb{E}_{s' \sim \tilde{P}_{\text{sim}}} [V_{\text{sim}}^\pi(s')] \right|.
\end{equation}

Adding and subtracting $\mathbb{E}_{s' \sim \tilde{P}_{\text{real}}} [V_{\text{sim}}^\pi(s')]$ inside the absolute value, and applying the triangle inequality:
\begin{equation}
\begin{aligned}
\Delta V(s) \leq\;& \gamma \mathbb{E}_{s' \sim \tilde{P}_{\text{real}}}
\left[ |V_{\text{real}}^\pi(s') - V_{\text{sim}}^\pi(s')| \right] + \gamma \\
&\left| \mathbb{E}_{s' \sim \tilde{P}_{\text{real}}} [V_{\text{sim}}^\pi(s')]
- \mathbb{E}_{s' \sim \tilde{P}_{\text{sim}}} [V_{\text{sim}}^\pi(s')] \right|.
\end{aligned}
\end{equation}

Let $\| \Delta V \|_\infty = \sup_s \Delta V(s)$ denote the maximum value discrepancy. Substituting the single-step error bound into the inequality:
\begin{equation}
    \| \Delta V \|_\infty \leq \gamma \| \Delta V \|_\infty + \gamma L_V \epsilon.
\end{equation}

Rearranging the terms to solve for $\| \Delta V \|_\infty$, we obtain the final bound to complete the proof:
\begin{equation}
    (1 - \gamma) \| \Delta V \|_\infty \leq \gamma L_V \epsilon \implies \| \Delta V \|_\infty \leq \frac{\gamma L_V}{1 - \gamma} \epsilon.
\end{equation}
\bibliographystyle{IEEEtran}
\bibliography{references}
\end{document}